# TBGC: Task-level Backbone-Oriented Gradient Clip for Multi-Task Foundation Model Learning

Zelun Zhang*,  Xue Pan
Huazhong University of Science and Technology
Wuhan, China
`zhangzelun123@gmail.com, panxue826@163.com`

## Abstract

*The AllInOne training paradigm squeezes a wide range of tasks into a unified model in a multi-task learning manner. However, optimization in multi-task learning is more challenge than single-task learning, as the gradient norm from different tasks may vary greatly, making the backbone overly biased towards one specific task. To address this issue, we propose the task-level backbone-oriented gradient clip paradigm, compared with the vanilla gradient clip method, it has two points of emphasis:1) gradient clip is performed independently for each task. 2) backbone gradients generated from each task are rescaled to the same norm scale. Based on the experimental results, we argue that the task-level backbone-oriented gradient clip paradigm can relieve the gradient bias problem to some extent. We also propose a novel multi-branch data augmentation strategy where conflict augmentations are placed in different branches. Our approach has been shown to be effective and finally achieve 1st place in the Leaderboard A and 2nd place in the Leaderboard B of the CVPR2023 Foundation Model Challenge. It's worth noting that instead of evaluating all three tasks(detection, segmentation and fine-grained classification) in Leaderboard A, the segmentation task is not evaluated in Leaderboard B, in which our team has a huge advantage[1].*

## 1. Introduction

As ChatGPT become the game changer in the natural language processing (NLP) field, foundation model, which is defined as "a model that is trained on broad data at scale and can be adapted to a wide range of downstream tasks", has become a research hotspot. The most generally used training paradigm of foundation model is Self-Supervision training at scale, such as Bert[1] in the NLP field, and SimCLR[2], MoCo [3] in the computer vision (CV) field, these self-supervision foundation model training paradigms are usually data hungry, requiring a huge amount of data, for example, Bert is pretrained on a 3.3 billion corpora, which is unpractical for fast industry deployment. Besides, the adaption cost for downstream tasks is also inescapable, which usually requires a from-scratch acquisition of the downstream-task datasets and a transfer learning (usually finetune) of the pretrained model parameters. Thus, the self-supervision training paradigm of foundation model has high requirements for data amount and computing power.

Due to the inefficiency and huge computation cost of the self-supervision pretraining, the AllInOne training paradigm is proposed to boost the foundation model learning. Unified Feature Optimization (UFO) [4] points out that some downstream tasks are related, it can be beneficial to train the foundation model in a multi-task and parameter-sharing manner, thus eliminate the requirement for huge amount of data for pretraining and reduce the adaption cost when transfer to downstream tasks.

In this paper, we adopt the AllInOne training paradigm to train a foundation model in a multi-task manner, acquiring a unified model that has the ability of detection, segmentation and fine-grained classification at the same time. During training, we find that the backbone gradient norm generated from different task may vary greatly, making the backbone overly biased towards one specific task, thus seriously impact the performance of other tasks. To address this problem, we propose the task-level backbone-oriented gradient clip paradigm, compared with the vanilla gradient clip method which perform gradient clip over the overall multi-task gradient, the task-level backbone-oriented gradient clip paradigm has two main differences: 1) gradient clip is performed independently over the gradient generated by each task. 2) the final backbone gradient generated from each task are rescaled to the same norm scale. Thus during training, each task has exact the sample influence on the backbone, avoiding the backbone gradient bias problem. Based on the experimental results, we argue that this novel gradient clip paradigm can relieve the gradient bias problem to some extend, achieving decent performance on the three tasks. Besides, we also noting the data augmentation conflict problem during training, i.e., the combined use of two specific data augmentation strategies may lead to a performance drop while using exclusively can steadily improve the performance, in terms of this issue, we propose the multi-branch data augmentation strategies, in which conflicting augmentation strategies are placed in different branches, thus the model can gain benefits from the originally conflicting strategies.

---
[1] * The first and the corresponding author

## 2. Method

### 2.1. Framework Overview

The track1 of CVPR2023 Foundation Model challenge requires training a model in the AllInOne multi-task manner, i.e., the unified model must have the ability of detection, semantic segmentation and fine-grained classification at the same time. The detection evaluation is conducted on the Tsinghua-Tencent100k dataset, taking mAP50 as metric. Semantic segmentation and fine-grained classification are evaluated on the BDD 100K dataset and Stanford Cars dataset respectively, using top1 accuracy and mIoU as evaluation metrics.

**Model Structure.** We adopt the common single-base but multiple-heads paradigm in terms of the model structure design, i.e., the model has a shared bottom (backbone) for feature extraction and each task has its own exclusive prediction head. We adopt the InternImage[5] series as backbone, Mask2Former[6] and DINO[7] are used for semantic segmentation head and detection head respectively, as for the classification head, we simply use a MLP with shortcut connection, i.e., the global average pooling feature from the last layer of backbone is firstly fed into a MLP, then the transformed feature output from the MLP and the backbone feature are added to obtain the final feature for fine-grained classification, besides, arcface loss[9] (with margin set to 0.4) is used.

**Training Process.** The training process design is crucial for multi-task foundation model learning, both computational cost and CUDA memory requirements are need to be taken into consideration. Given the computation graph of Pytorch takes nonnegligible CUDA memory during training, we argue that it's better to release the computation graph when iteration for one task is done. The training process we adopted is detailed in Algorithm 1.

By using this training process, in which each task's computation graph is released immediately after its iteration is done, the CUDA memory is greatly saved, we thus can use a much larger batchsize. The use of **T**ask-level **B**ackbone-oriented **G**radient **C**lip (TBGC) is also shown in Algorithm 1, which will be given a detailed description in the next section.

### 2.2. Task-level Backbone-oriented Gradient Clip

When training a foundation model in the AllInOne

---

**Algorithm 1**: Training Process with TBGC

**Input:** multi-task dataloader D, multi-task model M, epochs N, iteration overall gradient recorder R

1: initialize parameters in M
2: **for** i ← 1 to N **do**
3:    **for** multitask-data **in** D **do**
4:       zero initialize R
5:       **for** task-data, task-label **in** multitask-data **do**
6:          task-out = M(task-data)
7:          loss = loss(task-out, task-label)
8:          loss.backward() # release computation graph
9:          TBGC for M.grads
10:         accumulate clipped M.grads to R
11:         zero initialize M.grads
12:       **end**
13:       update M.grads with R
14:       M.step()
15:    **end**
16: **end**

---

multi-task manner, we find that the gradient norm from different task may vary greatly (as shown in Figure 1), thus leading the backbone to be overly biased to one specific task. Here we refer to "backbone gradient norm" as the l2 norm of the gradients on the backbone parameters. As shown in Figure 1, during training, the detection task has the largest backbone gradient norm, much larger than the counterparts of segmentation and classification. Besides, segmentation has a larger backbone gradient norm than classification. This explains why the detection task has a much better performance than the others and classification has the worst performance when the vanilla gradient clip method is adopted.

**Vanilla Gradient Clip.** Gradient clip is a common used strategy when training the transformer-series backbones. However, when it comes to the multi-task scenario discussed above, the vanilla gradient clip method suffers from the gradient norm bias problem, i.e., after the vanilla gradient clip, the clipped gradient still bias to the task which originally has the largest gradient norm. In summary, the vanilla gradient clip method under multi-task scenario can be formulated as:

$$Grad_{clipped} = \frac{Grad_{det} + Grad_{seg} + Grad_{cls}}{\|Grad_{det} + Grad_{seg} + Grad_{cls}\|_2} \times S \quad (1)$$

, $S$ stands for max norm, $Grad_{det}$, $Grad_{seg}$ and $Grad_{cls}$ stands for the gradients generated from each task in a multi-task iteration. $Grad_{clipped}$ stands for the final gradient used to update model parameters.

**TBGC.** The proposed Task-level Backbone-oriented Gradient Clip paradigm has two steps. Firstly, the vanilla gradient clip is conducted on each task independently.

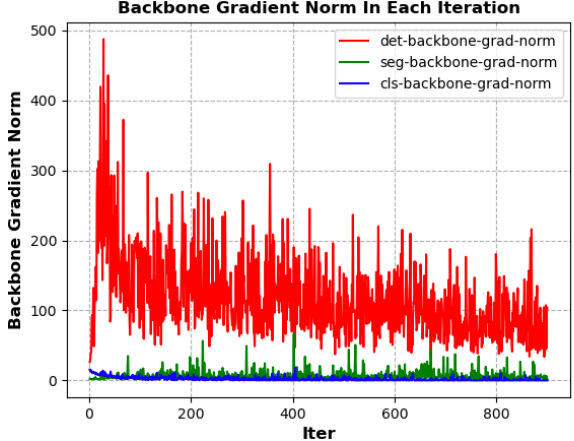

Figure 1. The backbone gradient norm of each task during training, samples are collected from 900 iterations from 100 epochs.

Secondly, the backbone gradient norm of each task is rescaled to the same norm scale. Thus each task has the exact same influence on the backbone parameters. In summary, the task-level backbone-oriented gradient clip paradigm can be formulated as:

$$Grad_{det}^1 = \frac{Grad_{det}}{\|Grad_{det}\|_2} \quad (2)$$

$$Grad_{det}^{clipped} = Grad_{det}^1 \times \frac{S}{backbone\_grad\_norm(Grad_{det}^1)} \quad (3)$$

, where $Grad_{det}$ is the original gradient generated from the detection task, $S$ stands for max norm, $backbone\_grad\_norm$ stands for a function that calculate the backbone gradient norm given the all model parameter gradients. It is worth noting that the scaling step (formula 3) is backbone-oriented, i.e., the backbone gradient norm will have the exact value of $S$ after TBGC. We obtain the segmentation and classification gradients in a similar way, finally, the overall gradient used to update the multi-task model is formulated as:

$$Grad_{clipped} = Grad_{det}^{clipped} + Grad_{seg}^{clipped} + Grad_{cls}^{clipped} \quad (4)$$

, where $Grad_{clipped}$ stands for the final overall gradient, $Grad_{seg}^{clipped}$ and $Grad_{cls}^{clipped}$ stands for the gradients generated by TBGC for segmentation and classification respectively. Compared with the vanilla gradient clip method, TBGC eliminates the gradient norm bias problem and make sure that every task has the exact save level of influence on the backbone parameters. Experiments in the next section will demonstrate the effectiveness of TBGC.

### 2.3. Multi-Branch Augmentation Paradigm

Data augmentation is a commonly used strategy to boost model performance. However, during training we find that the combined use of several augmentations might lead to a performance drop while using exclusively will improve the performance steadily. We argue that it is because some augmentation strategies are so strong that they cause a huge change to the original data distribution, and a combined use of them will make the distribution of training data differ hugely from the counterpart of testing data, thus harm the model performance.

Here we propose the multi-branch data augmentation paradigm. In this paradigm, each branch can have one and only one strong augmentation, such as Mosaic and Autoaugment. RandomChoice is adopted to combine these branches. During training, the data flow can only pass through one branch, only transformed by one strong augmentation, thus the model can gain benefits from multiple strong augmentations and can avoid the train-test inconsistency problem. Besides, we can introduce curriculum learning to the muti-branch paradigm, i.e., at the early stage of training, harder branch may have a larger probability while at end stage the easier branch may dominate. Table 1 shows the multi-branch augmentation strategies we used.

| detection | | segmentation | |
|---|---|---|---|
| branch1 | branch2 | branch1 | branch2 |
| *MultiScale* | *Mosaic* | *MultiScale* | *Mosaic* |
| *Hflip* | *MixUp* | *RandomCrop* | *RandomCrop* |
| *AutoAugment* | *Hflip* | *Rotate* | *Hflip* |
| *Noise* | *Noise* | *Noise* | *Noise* |

Table 1. the multi-branch data augmentation strategies used for CVPR2023 foundation model challenge.

Only detection and segmentation adopt the multi-branch augmentation paradigm, we use common augmentation paradigm for classification.

## 3. Experiments

**Hyperparameter Settings.** To evaluate our method, we conduct ablation experiments for the task-level backbone-oriented gradient clip paradigm and multi-branch augmentation strategy respectively. Unless specifically stated, we adopt the following hyperparameter settings. The training batchsize for detection, segmentation and fine-grained classification is 2, 2 and 8 per GPU respectively. The learning rate is 0.0001, weight decay is set to 1e-4. The cosine learning rate decay scheduler with warm up is adopted, specifically, the first 5 epochs is used for learning rate warm up, with a warm up ratio of 0.001, the model is trained for 100 epochs. We set the max norm hyper parameter to be 0.1 for both vanilla gradient clip and task-level backbone-oriented gradient clip. Besides, we adopt the adamW optimizer, the backbone is initialized with the pretrained parameters from ImageNet, and a differential learning rate is adopted, the backbone learning rate is 0.1 times of the base learning rate.

| Method | Overall | Det | Seg | Cls |
|---|---|---|---|---|
| *vanilla* | 77.50 | 86.78 | 56.67 | 89.06 |
| *TBGC\** | 79.82 | 87.60 | 58.85 | 92.99 |
| *TBGC* | **80.45** | **87.88** | **60.17** | **93.31** |

Table 2. The ablation study for vanilla gradient clip and TBGC. Det, Seg and Cls stands for detection, segmentation and classification respectively, they are evaluated with mAP50, mIoU and top-1 accuracy. * means TBGC without backbone-oriented norm rescaling.

**Data Augmentation.** Unless specifically stated, we adopt a plain data augmentation strategy. For detection, we use multi-scale training with the long edge set to 608 and the short edge randomly sampled from [480, 608], horizontal flip is also adopted. For segmentation, we use random resize with ratio range sample from [0.5, 2.0], horizontal flip and color distort is adopted. For classification, we directly resize the image to (448, 448), and then go through random erase and horizontal flip.

**Experiment Metric.** In our experiment, the detection task is evaluated with mAP50, the classification task is evaluated with top-1 accuracy and the segmentation task is evaluated with mIoU. Finally, metrics from the three tasks were averaged to get the overall metric.

### 3.1. Results of TBGC

Table 2 shows the comparison between different gradient clip methods, "*vanilla*" stands for the vanilla gradient clip method, i.e., gradient clip is conducted directly on the overall multi-task gradient. "*TBGC\**" stands for TBGC without backbone-oriented norm rescaling. The result shows the effectiveness of TBGC, without bells and whistles, the fine-grained classification task and segmentation task gains an increment of 4.25% and 3.5% respectively, without harming the performance of detection.

Even though performing vanilla gradient clip for each task independently is enough to relieve the gradient norm bias problem, given the fact that each task's exclusive prediction head has different amount of parameters, vanilla gradient clip cannot assure the task-level backbone gradient norm is all the same. Table 2 compare the performance between TBGC and TBGC*, TBGC* is fairly enough to give a decent performance, however, TBGC boost the overall performance further, which shows the importance of each task having the exact same influence on the backbone parameters.

| Method | Overall | Det | Seg | Cls |
|---|---|---|---|---|
| *parallel* | 85.79 | 94.04 | 68.29 | 95.03 |
| *MultiBranch* | **86.32** | **95.08** | **68.67** | **95.2** |

Table 3. The comparison between parallel augmentation strategy and multi-branch augmentation strategy.

### 3.2. Results of Multi-Branch Augmentation

Table 3 shows the comparison between parallel augmentation and multi-branch augmentation, the augmentations described in section 2.3 is adopted. Compared with the parallel strategy, the multi-branch counterpart achieves a 0.53 performance gain, which demonstrates that it is better to place strong augmentations in different branches.

## 4. Conclusion

In this paper, a novel gradient clip and data augmentation paradigm is proposed, experiments show the effectiveness of our proposed methods.